\documentclass{article}

\usepackage[preprint]{nips_2018}

\usepackage{pslatex}

\usepackage[pdftex]{graphicx}
\usepackage{comment}
\usepackage{amsmath}
\usepackage{amssymb}
\usepackage{xcolor}
\usepackage{hyperref}

\title{Learning robust visual representations\\using data augmentation invariance}
 
\author{{\large \bf Alex Hern\'andez-Garc\'ia (ahernandez@uos.de)} \\
  Institute of Cognitive Science, University of Osnabr\"uck\\
  27 Wachsbleiche, 49090 Osnabr\"uck, Germany
  \AND {\large \bf Peter K\"onig\textsuperscript{*} (pkoenig@uos.de)} \\
  Institute of Cognitive Science, University of Osnabr\"uck\\
  27 Wachsbleiche, 49090 Osnabr\"uck, Germany
  \AND {\large \bf Tim C. Kietzmann\textsuperscript{*} (tim.kietzmann@mrc-cbu.cam.ac.uk)} \\
  MRC Cognition and Brain Sciences Unit, University of Cambridge\\
  15 Chaucer Road, CB2 7EF Cambridge, UK
  \\
  \\\textsuperscript{*} shared senior authorship}

\begin{document}

\maketitle

\begin{abstract}
Deep convolutional neural networks trained for image object categorization have shown remarkable similarities with representations found across the primate ventral visual stream. Yet, artificial and biological networks still exhibit important differences. Here we investigate one such property: increasing invariance to identity-preserving image transformations found along the ventral stream. Despite theoretical evidence that invariance should emerge naturally from the optimization process, we present empirical evidence that the activations of convolutional neural networks trained for object categorization are not robust to identity-preserving image transformations commonly used in data augmentation. As a solution, we propose \textit{data augmentation invariance}, an unsupervised learning objective which improves the robustness of the learned representations by promoting the similarity between the activations of augmented image samples. Our results show that this approach is a simple, yet effective and efficient (10 \% increase in training time) way of increasing the invariance of the models while obtaining similar categorization performance.
\end{abstract}

\begin{quote}
\end{quote}

\section{Introduction}
\label{sec:intro}

Deep artificial neural networks (DNNs) have borrowed much inspiration from neuroscience and are, at the same time, the current best model class for predicting neural responses across the visual system in the brain \citep{kietzmann2017computneuro, kubilius2018cornet}. Yet, despite consensus about the benefits of a closer integration of deep learning and neuroscience \citep{bengio2015biologicallyplausible, marblestone2016dlandneuroscience}, important differences remain.

Here, we investigate a representational property that is well established in the neuroscience literature on the primate visual system: the increasing robustness of neural responses to identity-preserving image transformations. While early areas of the ventral stream are strongly affected by variation in e.g. object size, position or illumination, later levels of processing are increasingly robust to such changes \citep{isik2013dynamics}. The cascaded achievement of invariance to such identity-preserving transformations has been proposed as a key mechanisms for obtaining robust object recognition \citep{dicarlo2007untangling, tacchetti2018invariance}.

Learning such invariant representations has been a desired objective since the early days of artificial neural networks \citep{simard1992tangentprop}. Accordingly, a myriad of techniques have been proposed to attempt to achieve tolerance to different types of transformations (see \citet{cohen2016groupequivcnns} for a review). Interestingly, recent theoretical work has shown that invariance to ``nuisance factors'' should naturally emerge from the optimization process \citep{achille2018emergence}.

Nevertheless, DNNs are still not robust to identity-preserving transformations, including simple image translations \citep{zhang2019convolutions}, or more elaborate adversarial attacks \citep{szegedy2013adversarial}, in which small changes, imperceptible to the human brain, can alter the classification output of the network. In this regard, there is growing evidence that DNNs may exploit highly discriminative features that do not match human perception \citep{ilyas2019advfeatures}. Extending this line of research, we use image perturbations using the data augmentation framework \citep{hernandez2018data} to show that DNNs, despite being trained on augmented data, are not sufficiently robust to such transformations. 

Inspired by the increasing invariance observed along the primate ventral visual stream, we subsequently propose a simple, yet effective and efficient mechanism to improve the robustness of the representations: we include an additional term in the objective function that encourages the similarity between augmented examples within each batch.

\section{Methods}
\label{sec:methods}

This section presents the procedure to empirically measure the invariance of the representations of a convolutional neural network and our proposal to improve the invariance.

\subsection{Model, data and training parameters}
\label{sec:model}

As a test bed for our hypotheses and proposal we use the all convolutional network, All-CNN \citep{springenberg2014allcnn}, a well-known architecture which achieves good performance in spite of being much shallower than other architectures, and thus faster to train and more convenient for the analysis. It consists of 9 convolutional layers, with a total of 1.3 million parameters. Our model is identical to All-CNN-C in the original paper, except that we remove the explicit regularizers---weight decay and dropout---following the conclusions from \citet{hernandez2018data}. We also keep the original training hyperparameters: 350 epochs, initial learning rate of 0.01 and batch size of 128.

We train on the highly benchmarked data set for object recognition CIFAR-10 \citep{krizhevsky2009cifar} and apply heavier data augmentation than in the original paper. Specifically, we use the \textit{heavier} training and evaluation scheme described by \citet{hernandez2018data}, which includes random affine transformations and contrast and brightness adjustment.

\subsection{Evaluation of invariance}
\label{sec:eval}

To measure the invariance of the learned features under the influence of identity-preserving image transformations we compare the activations of a given image with the activations of a data augmented version of the same image. 

Consider the activations of an input image $x$ at layer $l$ of a neural network, which can be described by a function $f^{(l)}(x) \in \mathbb{R}^{D^{(l)}}$. We can define the distance between the activations of two input images $x_{i}$ and $x_{j}$ by their mean square difference:

\begin{equation}
\label{eq:mse}
 d^{(l)}(x_{i}, x_{j}) = \frac{1}{D^{(l)}}\sum_{k=1}^{D^{(l)}}(f_{k}^{(l)}(x_{i}) - f_{k}^{(l)}(x_{j}))^2
\end{equation}

Following this, we compute the mean squared difference between every $f^{(l)}(x_i)$ and a random transformation of $x_i$, that is $d^{(l)}(x_{i}, G(x_{i}))$. In this case, we define $G(x)$ as the data augmentation scheme that can take any of the extreme values of each transformation in the \textit{heavier} scheme, after halving the parameter ranges. This is to ensure the same level of augmentation in all comparisons, while preventing too extreme transformations.

The assessment of the similarity between the activations of an image $x_i$ and of its augmented versions $G(x_{i})$ was normalised by the similarity with the other, different images, reminiscent of an image identification problem. We define the invariance score $\sigma_{i}^{(l)}$ of the transformation $G(x_{i})$ at layer $l$ of a model, with respect to a data set of size $N$, as follows::

\begin{equation}
\label{eq:invariance}
 \sigma_{i}^{(l)} = 1 - \frac{d^{(l)}(x_{i}, G(x_{i}))}{\frac{1}{N}\sum_{j=1}^{N}d^{(l)}(x_{i}, x_{j})}
\end{equation}

\begin{figure*}[ht]
  \begin{center}
    \includegraphics[width = \linewidth]{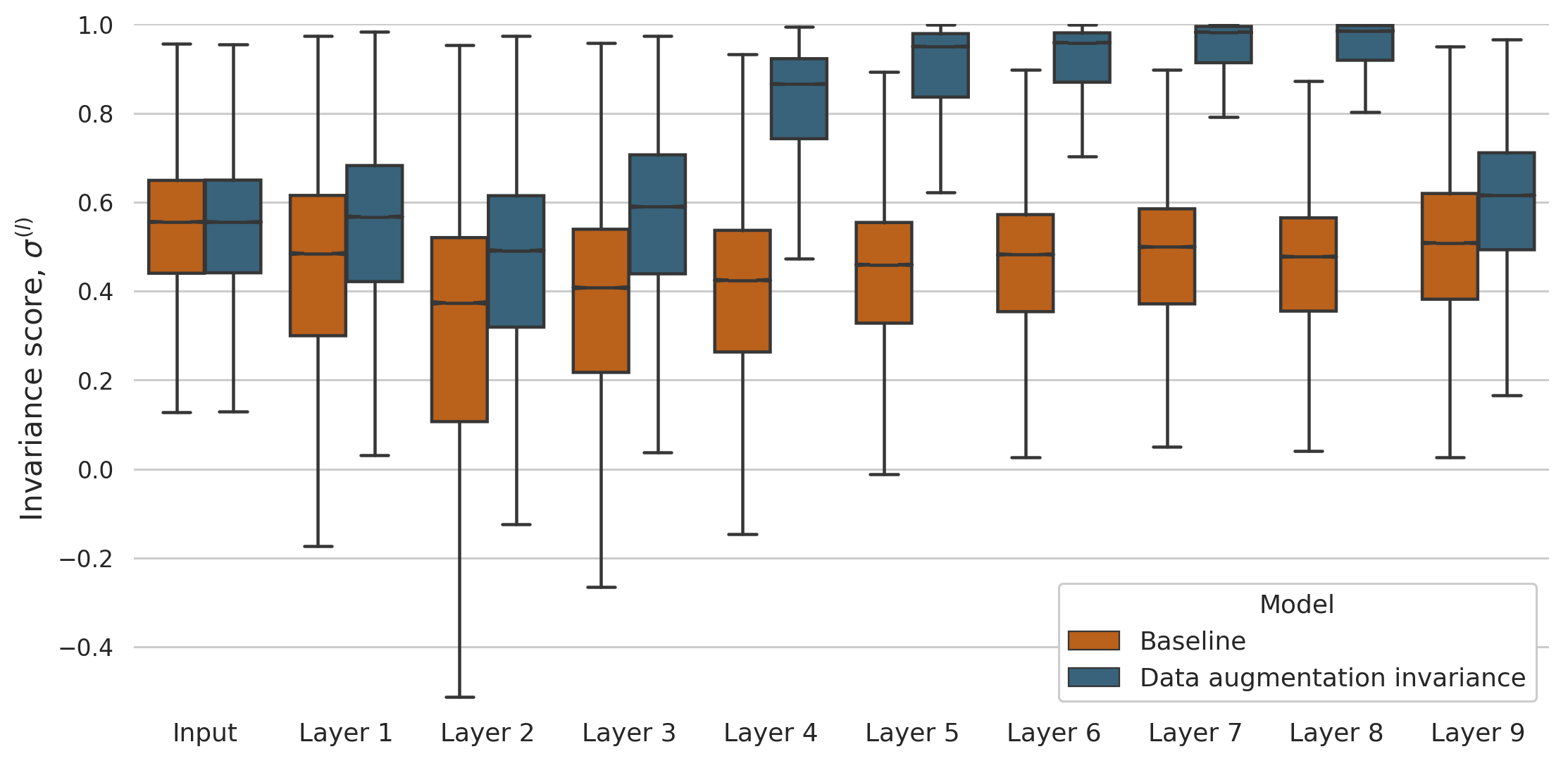}
  \end{center}
  \caption{Distributon of tnvariance score at each layer of the baseline model and the model trained data augmentation invariance.}
  \label{fig:invariance}
\end{figure*}

The invariance $\sigma_{i}^{(l)}$ takes the maximum value of 1 if the activations of $x_{i}$ and its transformed version $G(x_{i})$ are identical. To assess the overall invariance of a model post training, we calculate $\sigma_{i}^{(l)}$ for the 10,000 test images of CIFAR-10, with respect to five different random transformations. In Figure~\ref{fig:invariance} we show the distribution of $\sigma_{i}^{(l)}$ at each layer of All-CNN-C.

\subsection{Data augmentation invariance}
\label{sec:daug_invariance}

Most CNNs trained for object categorization are optimized through mini-batch gradient descent (SGD), that is the weights are updated iteratively by computing the loss of a batch $\mathcal{B}$ of examples, instead of the whole data set at once. The models are typically trained for a number of \textit{epochs}, $E$, which is a whole pass through the entire training data set of size $N$. That is, the weights are updated $K=\frac{N}{|\mathcal{B}|}$ times each epoch.

Data augmentation introduces variability into the process by performing a different, stochastic transformation of the data every time an example is fed into the network. However, with standard data augmentation, the model has no information about the \textit{identity} of the images, that is, that different augmented examples, seen at different epochs, separated by $\frac{N}{|\mathcal{B}|}$ iterations on average, correspond to the same seed data point. We believe this information may be valuable and useful to learn better representations in a self-supervised manner. For example, the high temporal correlation of the stimuli that reach the visual cortex may play a crucial role in the creation of robust connections \citep{wyss2006temporalstability}.

In order to make use of this information in an unsupervised way, we propose to perform data augmentation within the batches by constructing the batches to include $M$ transformations of each example (see \citet{hoffer2019batchaugmentation} for a similar idea). Additionally, we propose to modify the loss function to include an additional term that accounts for the invariance of the feature maps across multiple image samples. Considering the difference between the activations at layer $l$ of two images, $d^{(l)}(x_{i}, x_{j})$, defined in Equation~\ref{eq:mse}, we define the data augmentation invariance loss at layer $l$ for a given batch $\mathcal{B}$ as follows:

\begin{equation}
 \mathcal{L}_{inv}^{(l)} = \frac{\sum_{k}\frac{1}{|\mathcal{S}_{k}|^2}\sum_{x_i, x_j \in \mathcal{S}_{k}}d^{(l)}(x_{i}, x_{j})}{\frac{1}{|\mathcal{B}|^2}\sum_{x_i, x_j \in \mathcal{B}}d^{(l)}(x_{i}, x_{j})}
\end{equation}

where $\mathcal{S}_{k}$ is the set of samples in the batch $\mathcal{B}$ that are augmented versions of the same seed sample $x_k$. This loss term intuitively represents the average difference of the activations between the sample pairs that correspond to the same source image, relative to the average difference of all pairs. A convenient property of this definition is that $\mathcal{L}_{inv}$ does not depend on the batch size nor the number of in-batch augmentations $M=|\mathcal{S}_{k}|$. Furthermore, it can be efficiently implemented using matrix operations.

Since we want to achieve image invariance at $L$ layers of the network and jointly train for object recognition, we define the total loss as follows:

\begin{equation}
 \mathcal{L} = (1 - \alpha)\mathcal{L}_{obj} + \sum_{l=1}^{L}\alpha^{(l)}\mathcal{L}_{inv}^{(l)}
\end{equation}

where $\sum_{l=1}^{L}\alpha^{(l)} = \alpha$ and $\mathcal{L}_{obj}$ is the loss associated with the object recognition objective, typically the cross-entropy between the object labels and the output of a softmax layer. All the results we report in this paper have been obtained by setting $\alpha=0.1$ and distributing the coefficients across the layers according to an exponential law, such that $\alpha^{(l=L)}= 10\alpha^{(l=1)}$. This aims at simulating a probable response along the ventral visual stream, where higher regions are more invariant than the early visual cortex\footnote{It is beyond the scope of this paper to analyze the sensitivity of the hyperparameters $\alpha^{(l)}$, but we have not observed a significant impact in the classification performance by using other distributions.}.

\section{Results}
\label{sec:results}

One of the contributions of this paper is to empirically test in how far convolutional neural networks produce invariant representations under the influence of identity-preserving transformations of the input images. Figure~\ref{fig:invariance} shows the invariance scores, as defined in Equation~\ref{eq:invariance}, across network layers.

Despite the presence of data augmentation during training, which implies that the network may learn augmentation-invariant transformations, the representations of the baseline model (red boxes) do not increase in invariance beyond the pixel space. As a solution, we have proposed a simple, unsupervised modification of the loss function to encourage the learning of data augmentation-invariant features. As can be seen in Figure~\ref{fig:invariance} (blue boxes), our data augmentation mechanism pushed network representations to become increasingly more robust with network depth. One exception is the top, 'readout' layer, likely because the features are dominated by the categorization objective.

\begin{figure*}[ht]
  \begin{center}
    \includegraphics[width = \linewidth]{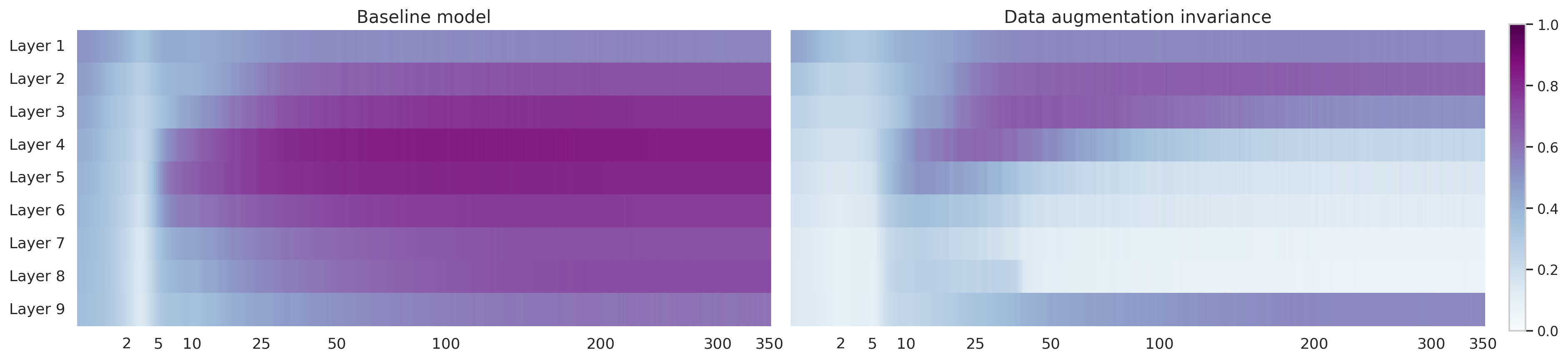}
  \end{center}
  \caption{Dynamics of the data augmentation invariance loss $\mathcal{L}_{inv}^{(l)}$ during training. The axis of abscissas (epochs) is scaled quadratically to better appreciate the dynamics at the first epochs. The same random initialization was used for both models.}
  \label{fig:dynamics}
\end{figure*}

In order to better understand the effect of the data augmentation invariance, we plotted the learning dynamics of the invariance loss at each layer. In Figure~\ref{fig:dynamics}, we can see that in the baseline model, the invariance loss keeps increasing over the course of training. In contrast, when the loss is added to the optimization objective, the loss drops for all but the last layer. Unexpectedly, the invariance loss increased during the first epochs and only then started to decrease. While further investigations are required, these two phases may correspond to the compression and diffusion phases proposed by \citet{shwartz2017bottleneck}.

In terms of efficiency, adding terms to the objective function implies an overhead of the computations. However, since the pairwise distances can be efficiently computed at each batch through matrix operations, the training time is only increased by about 10 \%. Finally, the improved invariance comes at no cost in the categorization performance, as the network trained with data augmentation invariance achieves similar classification performance to the baseline model---test accuracy baseline: 91.5 \%; test accuracy data augmentation invariance: 92.2 \%).

\section{Conclusions}
\label{sec:conclusions}

In this work we have empirically shown that the features learned by a prototypical convolutional neural networks are not invariant to identity-preserving image transformations despite being part of the training procedure. This property is fundamentally different to the primate ventral visual stream, where neural populations have been found to be increasingly robust to changes in view or lighting conditions of the same object \citep{dicarlo2007untangling}.

Taking inspiration from this property of the visual cortex, we have proposed an unsupervised objective to encourage learning more robust features, using data augmentation as the framework to perform identity-preserving transformations on the input data. We created mini-batches with $M$ augmented versions of each image and modified the loss function to maximize the similarity between the activations of the same seed images. 

Data augmentation invariance effectively produces more robust representations, unlike standard models optimized only for object categorization, at no cost in classification performance. Future work will investigate whether this increased robustness also allows for better modelling of neural data.

\section*{Acknowledgments}

This project has received funding from the European Union's Horizon 2020 research and innovation programme under the Marie Sklodowska-Curie grant agreement No 641805, from the Cambridge Commonwealth, European and International Trust, and the DFG.

\bibliographystyle{iclr2018_conference}
\bibliography{references}
\end{document}